\documentclass[12pt]{spieman}  
\usepackage{amsmath,amsfonts,amssymb}
\usepackage{graphicx}
\graphicspath{ {img/} }
\usepackage{soul}
\usepackage{easyReview}
\usepackage{setspace}
\usepackage{tocloft}
\usepackage{lineno}
\usepackage{threeparttable}
\usepackage{caption}
\usepackage{subcaption}
\usepackage{orcidlink}
\title{Mammographic Lesion Segmentation with Lightweight Models: A Comparative Study}

\author[a,*]{Helder Oliveira\,\orcidlink{0000-0003-4558-2844}}
\affil[a]{Independent Researcher, GnosisX, Calgary, AB, Canada}

\cftpagenumbersoff{figure}
\cftpagenumbersoff{table} 
\begin{document} 
\maketitle


\begin{abstract}
\textbf{Purpose}: Breast cancer is a leading cause of cancer-related mortality among women worldwide, with mammography as the primary screening tool. While deep learning models have demonstrated strong performance in lesion segmentation, most rely on computationally intensive architectures, which limit their use in resource-constrained environments. This study evaluates whether lightweight models can achieve clinically relevant segmentation performance under explicit computational constraints, namely, a CPU-only environment and limited memory availability.

\textbf{Approach}: Several lightweight architectures, including MobileNetV2, EfficientNet Lite, FPN, and Fast-SCNN, were compared against a U-Net baseline using the INbreast dataset under a 5-fold cross-validation protocol. Performance was assessed using Dice score, Intersection over Union (IoU), and Recall, alongside model complexity. The best-performing model was further evaluated on the DMID dataset to assess cross-dataset generalization.

\textbf{Results}: MobileNetV2 with Concurrent Spatial and Squeeze \& Excitation (SCSE) achieved the best performance, with a Dice score of 0.5766 while using approximately 75\% fewer parameters, highlighting a favorable accuracy–efficiency trade-off under constrained execution settings than U-Net. Cross-dataset evaluation on DMID showed reduced segmentation accuracy due to domain shift, with Dice and IoU decreasing, while Recall remained relatively stable.

\textbf{Conclusions}: Lightweight architectures can achieve competitive segmentation performance while significantly reducing computational cost. These findings demonstrate that lightweight architectures can meet practical deployment constraints while maintaining competitive segmentation performance, supporting their use in resource-limited CAD systems.
\end{abstract}


{
    \noindent \footnotesize\textbf{*}Helder Oliveira,  \linkable{helder@gnosisx.com}, \linkable{https://gnosisx.com}
}

\begin{spacing}{2}   
\section{Introduction}
\label{sect:intro}
Breast cancer is the most frequently diagnosed cancer among women worldwide and a leading cause of cancer-related mortality~\cite{who}. Early detection through screening programs significantly improves survival rates, with mammography being the most widely adopted imaging modality~\cite{oeffinger2015}. However, accurate interpretation of mammograms remains challenging due to subtle visual patterns, tissue overlap, and variability in lesion appearance~\cite{Elmore2003}.

Computer-aided detection (CAD) systems have been proposed to assist radiologists by improving sensitivity and reducing diagnostic variability~\cite{chotai2026}. In recent years, deep learning models, specifically Convolutional Neural Networks (CNNs), have demonstrated strong performance in mammographic analysis tasks, including classification, detection, and segmentation~\cite{Litjens2017, wu2020, loizidou2023}. Among these, segmentation plays a critical role, as it enables precise localization of suspicious regions, such as masses, calcifications, and architectural distortions, thereby supporting downstream tasks such as radiomics analysis and malignancy prediction~\cite{rezaei2021}.

The introduction of the U-Net architecture marked a significant milestone in medical image segmentation~\cite{ronnebergerUNet2015}. Its encoder--decoder structure with skip connections has since become a standard, with numerous variants achieving state-of-the-art performance across medical imaging domains~\cite{zhang2024, gonzalez2023, chen2024a}. In mammography, several studies have successfully applied U-Net-based or similar architectures for lesion segmentation~\cite{li2019a, prasetyo2024, hossain2022}.

Despite their effectiveness, these models are usually computationally intensive, often requiring Graphics Processing Units (GPUs) for both training and inference. This creates a significant barrier to deployment in resource-constrained environments, such as low- and middle-income countries, rural clinics, and point-of-care or edge devices, where access to high-performance computing infrastructure is limited~\cite{kelly2019, whoAI, ghadi2024, mashmool2026}. Furthermore, even in well-resourced settings, reliance on cloud-based inference introduces concerns related to latency, operational cost, and data privacy, particularly when sensitive medical data must be transmitted and processed remotely~\cite{mashmool2026, ghadi2024, rocha2024, kaissis2020}.

In contrast, the computer vision community has proposed several lightweight architectures, such as \emph{MobileNetV2}~\cite{sandler2018mobilenetv2}, \emph{EfficientNet}~\cite{tan2019efficientnet}, \emph{FPN}~\cite{lin2017fpn}, and \emph{Fast-SCNN}~\cite{poudel2019fastscnn}. These networks were designed to operate efficiently on devices with limited computational resources. While these models have been extensively evaluated in natural image tasks and real-time applications, their adoption in mammographic segmentation remains limited.

A critical gap in the literature is the lack of systematic evaluation of lightweight segmentation models in mammography. Most existing studies prioritize accuracy metrics such as the Dice score or Area Under the Curve (AUC), while neglecting computational cost and deployability, which are essential for real-world adoption~\cite{kelly2019}. To address these gaps, this study is guided by the following research questions:

\begin{itemize}
    \item Can lightweight segmentation models achieve competitive performance compared to standard U-Net-based architectures in mammographic lesion segmentation?
    \item Do lightweight models generalize across datasets with different acquisition characteristics?
    \item Which architectures offer the best accuracy–efficiency trade-off under explicit computational constraints (CPU-only environment and limited memory)?
\end{itemize}

Unlike prior work that prioritizes maximizing performance, this study aims to identify models that are practical to train and deploy under limited computational resources, thereby enabling broader accessibility of CAD systems. 

In this work, we explicitly define a practical deployment constraint by considering a CPU-only environment without GPU acceleration and limited memory availability. Within this setting, the goal is not to maximize segmentation accuracy, but to identify models that achieve competitive performance while remaining computationally feasible. This enables evaluation under realistic operational conditions rather than idealized high-performance environments.

\section{Related Works}

\subsection{Deep Learning in Mammography}

Deep learning has significantly advanced mammographic image analysis, particularly in classification and detection tasks. Early works demonstrated the potential of CNNs for breast cancer classification using a large-scale dataset of 1~M images~\cite{wu2020}. The study used the ResNet-22 architecture, which has over 6~M parameters, trained in an environment equipped with a GPU to achieve an AUC of $0.895$.

Another study using images from 1,912 patients proposed a two-step approach employing two deep learning networks to classify breast lesions~\cite{zheng2023}. The first step used the \emph{RefineNet} to do segmentation, followed by \emph{Xception} and Pyramid pooling to classify the regions segmented. The best Dice score reported was $0.888$ followed by an AUC of $0.947$. 

Comprehensive reviews highlight that while classification has reached a high level of maturity, segmentation remains comparatively less explored and presents additional challenges due to annotation complexity and class imbalance~\cite{loizidou2023, rani2025}.

\subsection{Medical Image Segmentation}
\label{secSegmentation}

\begin{figure}[!htb]
    \centering
    \includegraphics[scale=0.3]{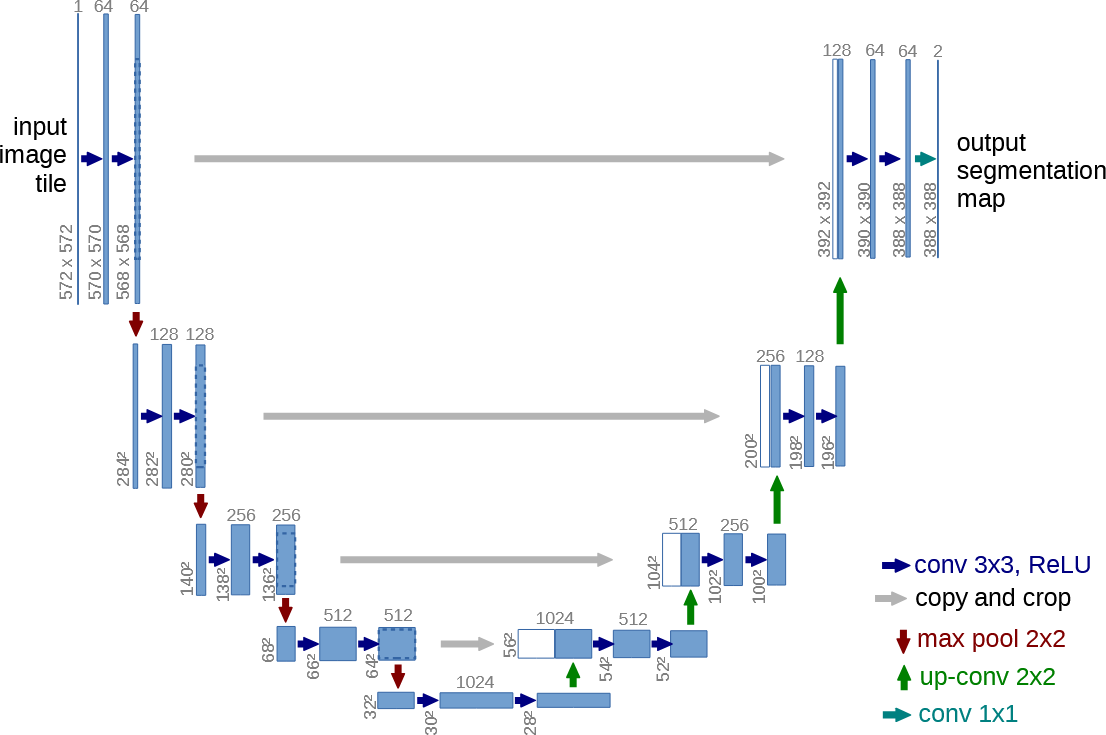}
    \caption{U-Net architecture~\cite{ronnebergerUNet2015}.}
    \label{figUnet}
\end{figure}

The U-Net~\cite{ronnebergerUNet2015} model, shown in Fig.~\ref{figUnet}, is presented here as a representative architecture due to its foundational role in medical image segmentation. Many modern lightweight segmentation models adopt similar encoder–decoder structures, often differing primarily in the choice of backbone network.

Extensions such as \emph{nnU-Net}~\cite{isensee2021nnunet} have further automated architecture design and achieved state-of-the-art results across multiple benchmarks and image modalities. Isensee~\emph{et al.}~(2021) evaluated the proposed architecture against 23 public datasets, demonstrating the generalization power of a single architecture in multiple domains.

A recent study~\cite{prasetyo2024} has evaluated the segmentation performance of \emph{U-Net} and two other variants, \emph{Res-UNet} and \emph{Att-UNet}, for the segmentation of breast ultrasound images. It was reported that, with a Dice score of $0.9799$ and Intersection of Union (IoU) of $0.9341$, \textit{U-Net} with attention (\emph{Att-UNet}) outperformed the other two architectures.

Chen~\emph{et al.}~(2024)\cite{chen2024a} proposed \emph{TransUNet}, a revised version of the original \emph{U-Net}. The new model employed Transformers to overcome \textit{U-Net}'s limitations of modeling long-range dependencies when dealing with 3D data, such as MRI and CT. Therefore, two \textit{TransUNet} versions were evaluated, one with Transformers only in the encoder (33.6M parameters), and another one with Transformers in both (41.4M parameters). The former reached an average Dice score of $87.63\%$, while the latter a Dice score of $88.39\%$. 

Another model proposed for medical image segmentation is \textit{LV-Unet}~\cite{jiang2025}. It uses \textit{MobileNet V3 - Large} as a backbone and fusible modules as a decoder. With some variations in the encoder-decoder layers, the number of parameters ranges between 0.8 and 2.8~M. After testing on five different datasets of medical images, the highest Dice and IoU scores are 0.9197 and 0.8531, respectively. 

Based on the \textit{Segment Anything Model} (SAM), the \textit{Localization Distillation-Enhanced Feature Prompting SAM} (LDFSAM) was proposed~\cite{zhao2026}. The model has two versions, one using YOLOv8n and another with YOLOv8x as backbone, totaling 274.5M parameters on average. After training and testing with four datasets of 3D medical images, the model reached Dice and IoU scores of 93.71 and 88.47, respectively. 

In mammography, Li~\emph{et al.}~(2019)~\cite{li2019a} proposed a deep learning framework for mass segmentation. Based on \textit{U-Net}, the new model used attention gates. The results reported, using the DDSM dataset, show higher performance compared to \textit{U-Net}, \textit{U-Net Attention}, and \textit{DenseNet}. 

Shen~\emph{et al.}~(2021)~\cite{shen2021} focused on the importance of having an interpretable classifier. Therefore, they proposed a deep learning model that first segments the image, looking for suspicious spots, and then performs the classification. It is mentioned that to keep resource consumption manageable, the authors decided to use \textit{ResNet-22} as part of the Global-aware Multiple Instance Classifier (\textit{GMIC}) for segmentation. The results showed that \textit{GMIC} and \textit{U-Net} had a Dice score of 0.325 and 0.504, respectively.

Microcalcification segmentation is challenging in mammography due to its subtle nature, which manifests as a few pixels. Hossain~(2022)~\cite{hossain2022}, using the DDSM dataset, proposed a U-Net variant capable of segmenting calcifications. The results show a Dice score of $97.8\%$, outperforming other models evaluated, such as \textit{DenseNet-II} and \textit{U-Net}.

Besides the remarkable advances, these approaches typically rely on large models and are trained on large datasets, demanding GPUs for training. It is worth mentioning that those models are trained and tested on the same dataset without exploring the generalization power across different datasets of the same image modality. Moreover, only a few are focused on mammography.

\subsubsection{Lightweight Architectures in the Literature}
%

Lightweight CNN architectures have been developed to reduce computational complexity while maintaining competitive performance. \emph{MobileNetV2}~\cite{sandler2018mobilenetv2} introduced inverted residual blocks and depthwise separable convolutions to reduce the number of parameters. Among the results reported, the \emph{MobileNetV2} creators reported an average IoU of 75\% in a validation set. The model had 2.11M parameters and approximately 2.75B Multiply-Adds operations (5.5B FLOPs).

Tan~\&~Le~(2019)~\cite{tan2019efficientnet} proposed a family of resource-efficient models called \emph{EfficientNet}. The models have a compound scaling method to balance network depth, width, and resolution. The results confirm how effective the model is, reaching an accuracy of 84.3\% on ImageNet with the model \emph{EfficientNet-B7} (66M of parameters and 37B FLOPs), which is 8.4x smaller than other \textit{ConvNets} evaluated.

For segmentation, \emph{Fast-SCNN}~\cite{poudel2019fastscnn} was specifically designed for real-time applications, achieving low latency and reduced memory usage. \textit{FPN}, for example, leverages multi-scale feature fusion through a pyramid structure, making it suitable for segmentation tasks requiring contextual information at different resolutions. \emph{Fast-SCNN} was developed to efficiently process high-resolution images. With only 1.11M parameters, \textit{Fast-SCNN} reached a mean IoU of 68\%.

Despite their advantages, the aforementioned architectures have been minimally explored in mammographic segmentation, and there is limited evidence of their performance in this domain.

\subsection{Gaps in the Literature}

Across the reviewed studies, several consistent gaps can be identified:
\begin{itemize}
    \item Lack of efficiency-focused evaluation: Most works prioritize accuracy while ignoring computational cost.
    \item Limited exploration of lightweight models: Architectures designed for edge deployment are rarely evaluated in mammography.
    \item Absence of deployment-oriented metrics: Few studies report the number of parameters or the number of operations.
    \item Poor cross-dataset validation: Models are often evaluated on a single dataset, limiting conclusions about generalization.
\end{itemize}

These limitations highlight the need for studies that explicitly consider both performance and deployability, which is the primary focus of this work.

\section{Materials and Methods}
In this section, the datasets used, the models' training and evaluation settings are discussed in detail. All experiments were developed by setting the necessary parameters for full reproducibility. Code, datasets, and configurations availability is provided in the OSF repository, which is provided in Sec.~\ref{secDataAvailability}.

\subsection{Mammography Datasets}
Deep learning-based segmentation~\cite{rani2025} studies in mammography commonly rely on publicly available datasets such as CBIS-DDSM~\cite{lee2017cbisddsm} and INbreast~\cite{moreira2012-InBreast-Dataset}. In some situations, private datasets are used, which makes it difficult to develop other studies. 

CBIS-DDSM~\cite{lee2017cbisddsm} is a curated version of the \emph{Digital Database for Screening Mammography} (DDSM) established in 2001~\cite{ddsm2001}. The newer version contains thousands of clinical images with many types of breast lesions carefully annotated by radiologists. Besides the great number of cases, CBIS-DDSM is composed of digitized film mammograms, which may result in lower image quality and annotation inconsistencies. 

The INbreast dataset provides high-resolution Full-field Digital Mammograms (FFDM), acquired with the \emph{Siemens} equipment \emph{MammoNovation}. The dataset is composed of 410 images, where 108 are normal tissue, and 302 are considered abnormal. Among the abnormal cases, there are 108 masses, 308 calcifications, and 3 architectural distortions. Note that a single image may contain multiple findings. Table~\ref{tblDataSets} shows a summary of the INbreast dataset.


\begin{table}[!htb]
    \centering
    \caption{Summary of the INbreast and DMID datasets.}
    \label{tblDataSets}
    \begin{threeparttable}
        \begin{tabular}{c|c|c|c}
            \hline\hline
            \rule[-1ex]{0pt}{3.5ex} \textbf{Dataset} & Normal & Abnormal & \textbf{Total images} \\
            \hline\hline
            \rule[-1ex]{0pt}{3.5ex} INbreast~\cite{moreira2012-InBreast-Dataset} & 108 & 302 & 410 \\
            \hline
            \rule[-1ex]{0pt}{3.5ex} DMID~\cite{oza2024-DMID-Dataset} & 241\tnote{*} & 269\tnote{*} & 510 \\
            \hline\hline
        \end{tabular}
        \begin{tablenotes}
            \footnotesize
            \item[*] These numbers correspond to the dataset analysis and differ from the reported values.
        \end{tablenotes}
    \end{threeparttable}
\end{table}

\begin{figure}[!htb]
     \centering
     \begin{subfigure}[b]{0.18\textwidth}
         \centering
         \includegraphics[scale=0.15]{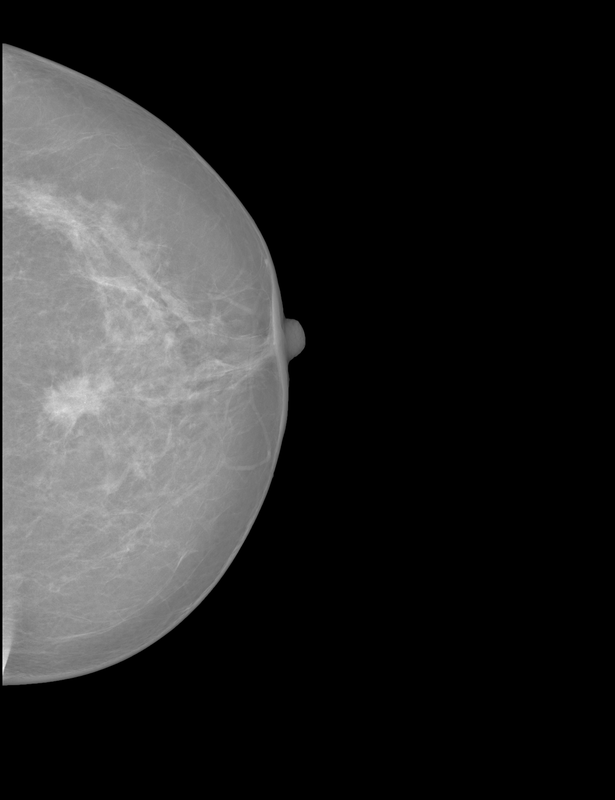}
         \caption{}
     \end{subfigure}
     \hfill
     \begin{subfigure}[b]{0.18\textwidth}
         \centering
         \includegraphics[scale=0.15]{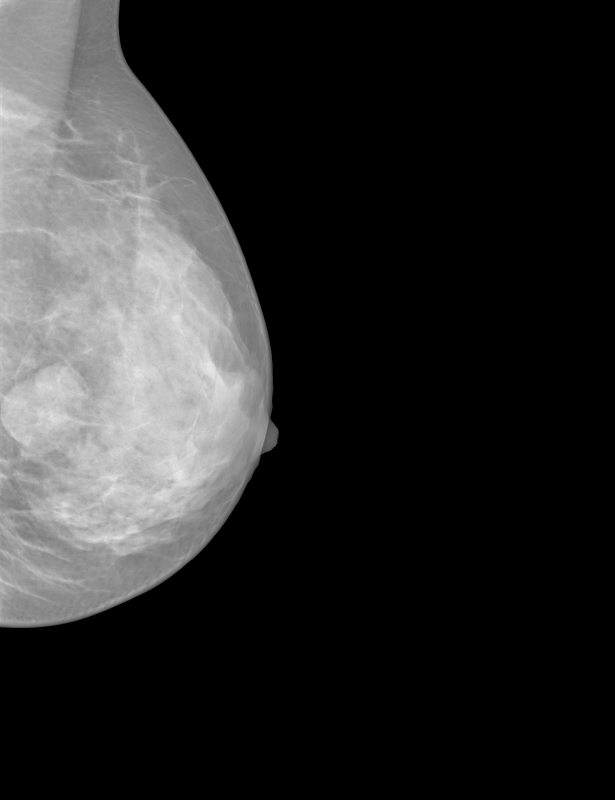}
         \caption{}
     \end{subfigure}
     \hfill
     \begin{subfigure}[b]{0.18\textwidth}
         \centering
         \includegraphics[scale=0.15]{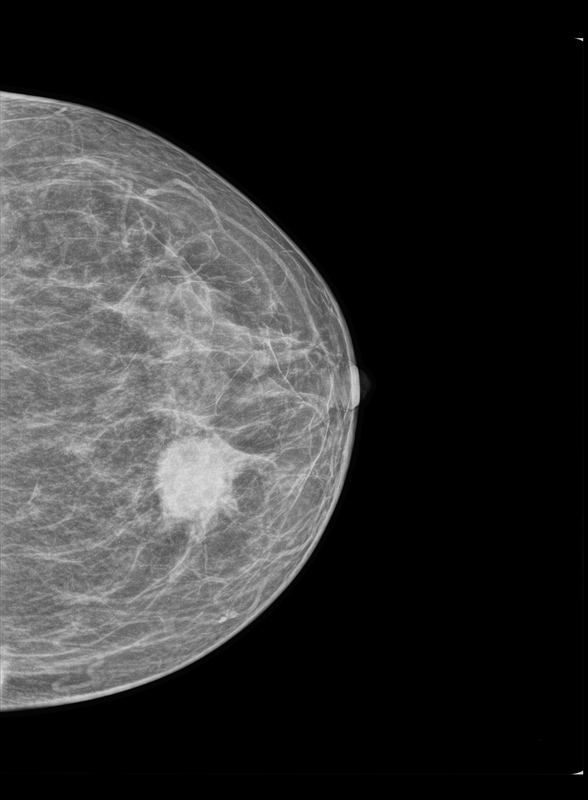}
         \caption{}
     \end{subfigure}
     \hfill
     \begin{subfigure}[b]{0.18\textwidth}
         \centering
         \includegraphics[scale=0.15]{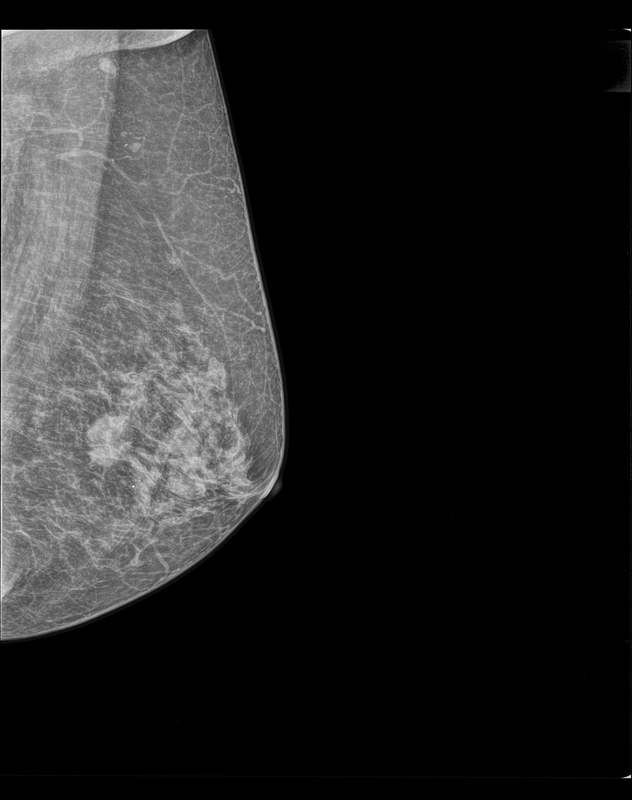}
         \caption{}
     \end{subfigure}
     \\
      \begin{subfigure}[b]{0.18\textwidth}
         \centering
         \includegraphics[scale=0.15]{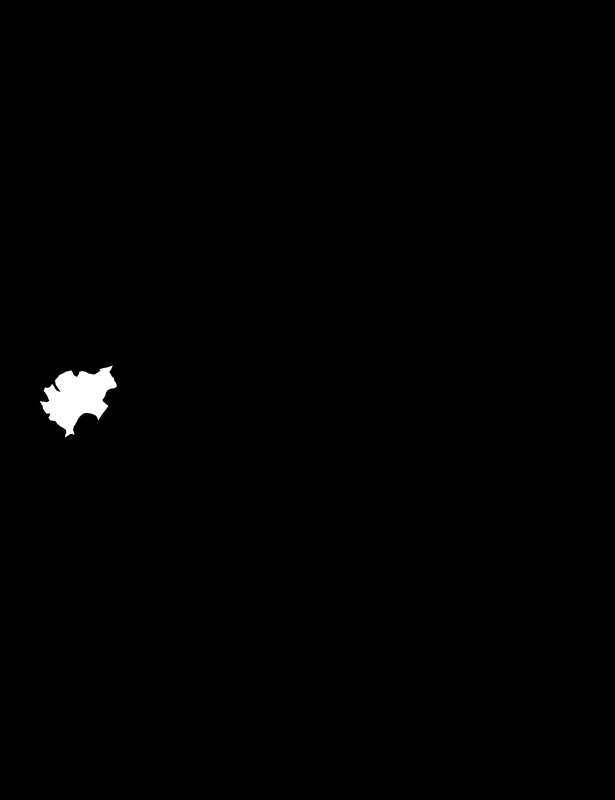}
         \caption{}
     \end{subfigure}
     \hfill
     \begin{subfigure}[b]{0.18\textwidth}
         \centering
         \includegraphics[scale=0.15]{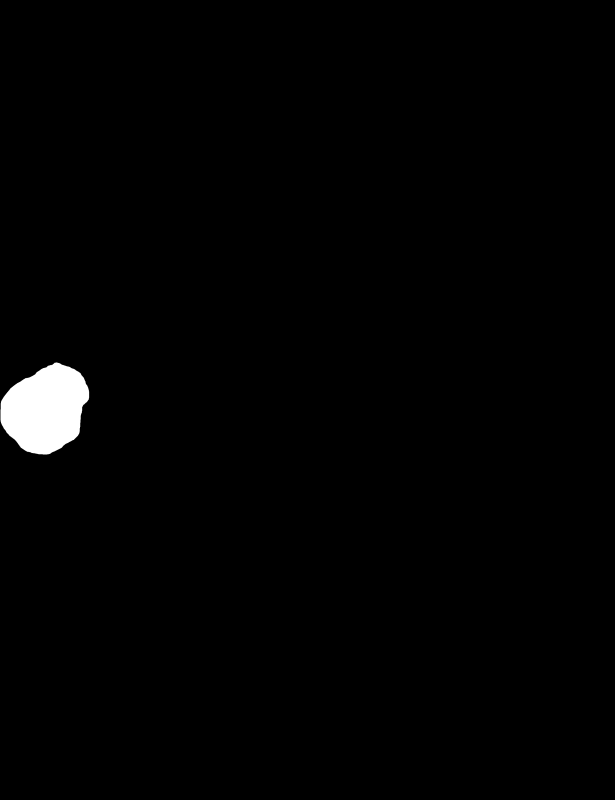}
         \caption{}
     \end{subfigure}
     \hfill
     \begin{subfigure}[b]{0.18\textwidth}
         \centering
         \includegraphics[scale=0.15]{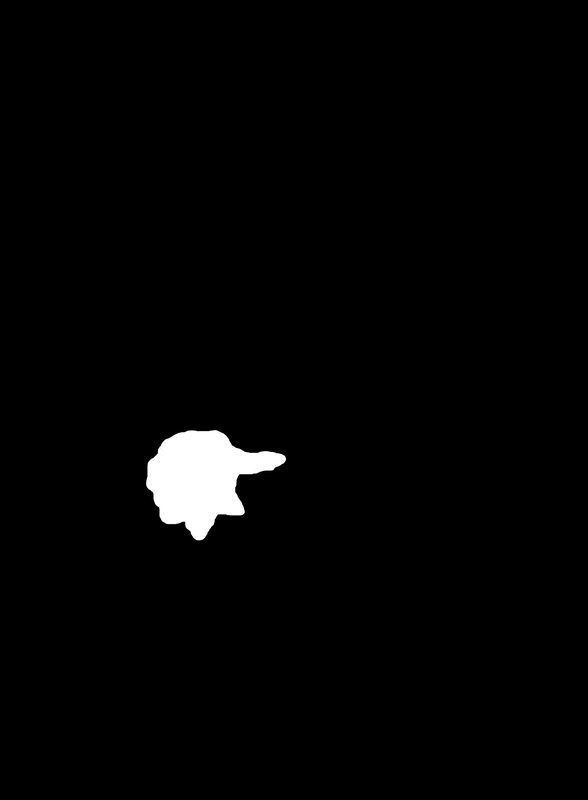}
         \caption{}
     \end{subfigure}
     \hfill
     \begin{subfigure}[b]{0.18\textwidth}
         \centering
         \includegraphics[scale=0.15]{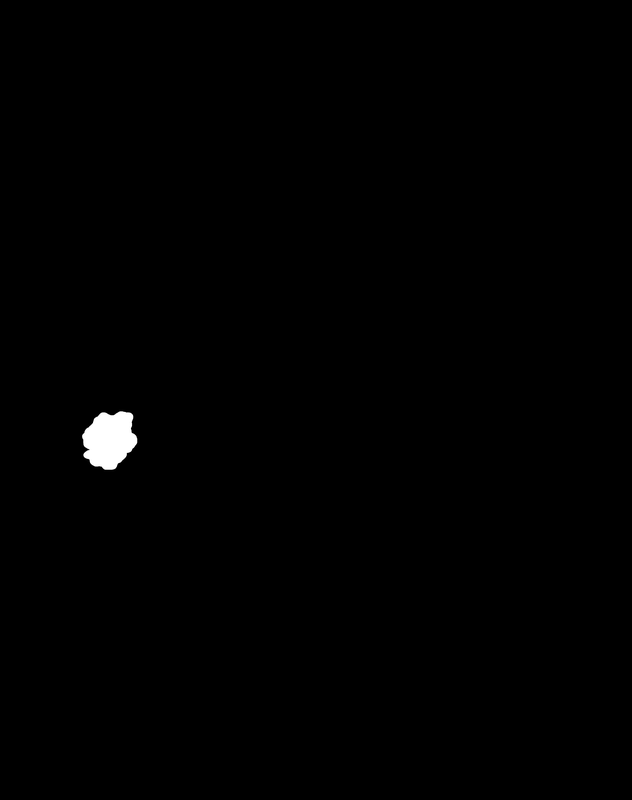}
         \caption{}
     \end{subfigure}
     \\
      \begin{subfigure}[b]{0.18\textwidth}
         \centering
         \includegraphics[scale=0.15]{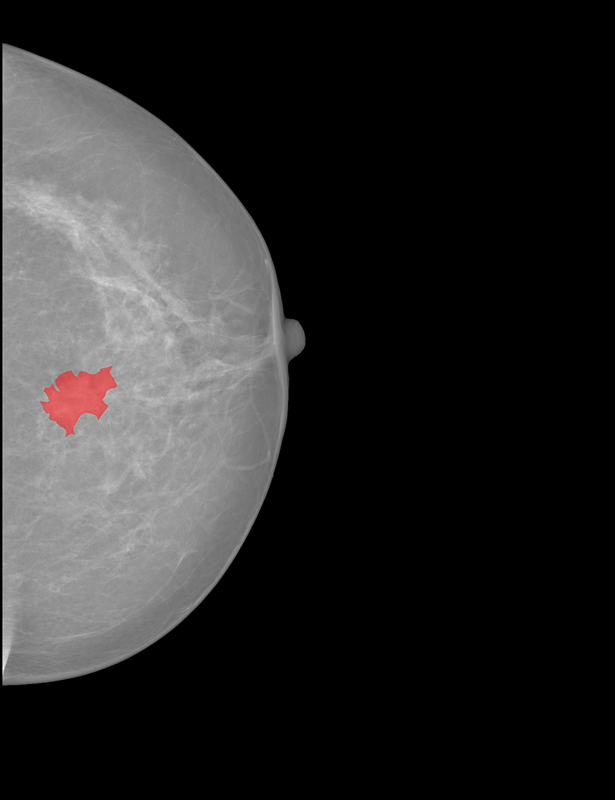}
         \caption{}
     \end{subfigure}
     \hfill
     \begin{subfigure}[b]{0.18\textwidth}
         \centering
         \includegraphics[scale=0.15]{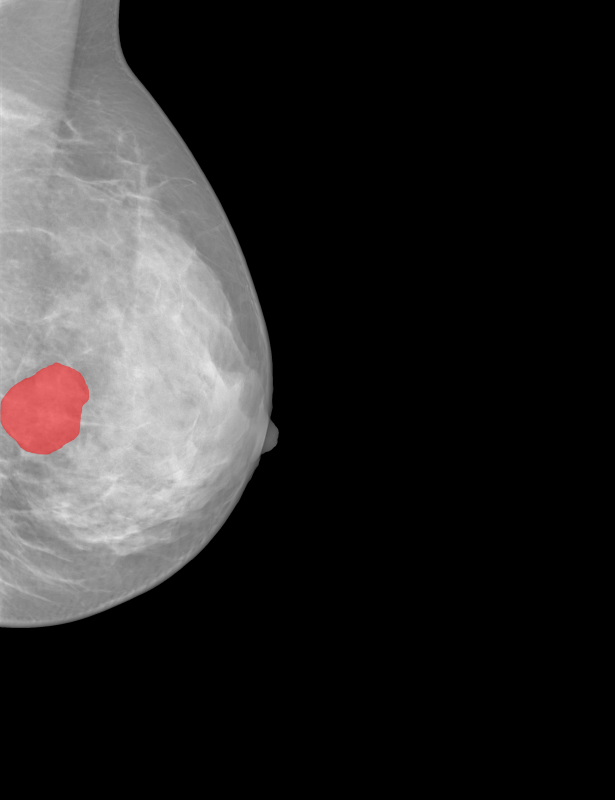}
         \caption{}
     \end{subfigure}
     \hfill
     \begin{subfigure}[b]{0.18\textwidth}
         \centering
         \includegraphics[scale=0.15]{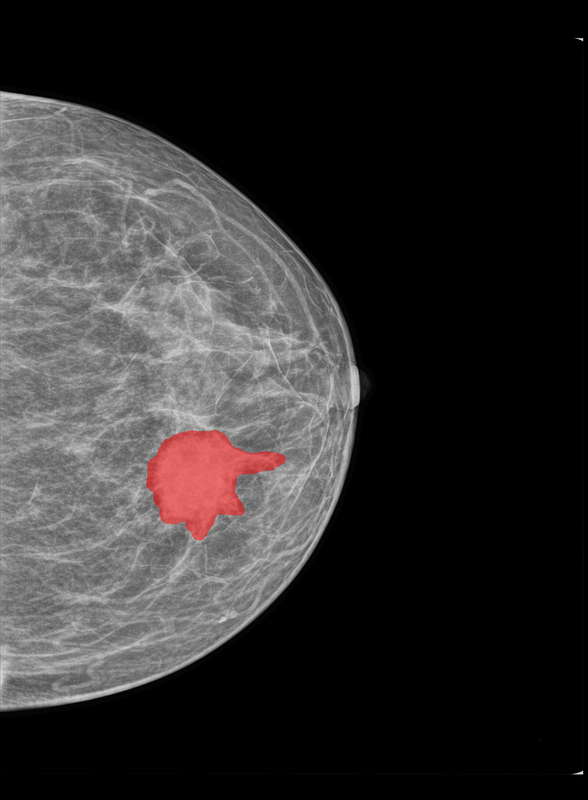}
         \caption{}
     \end{subfigure}
     \hfill
     \begin{subfigure}[b]{0.18\textwidth}
         \centering
         \includegraphics[scale=0.15]{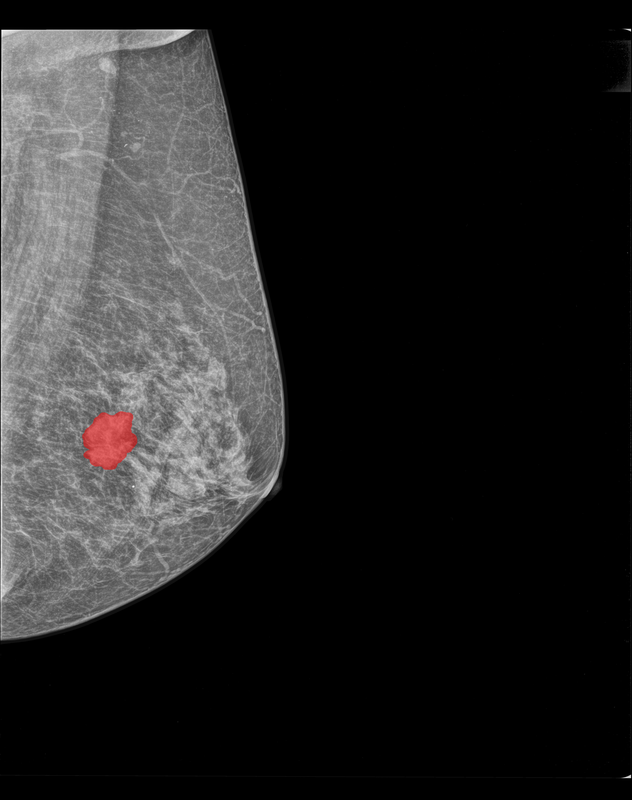}
         \caption{}
     \end{subfigure}
        \caption{Sample images from (a-b) INbreast and (c-d) DMID datasets. Each column shows the original image, the ground truth annotation, and the corresponding mask overlaid on the image.}
        \label{imgDatasetSamples}
\end{figure}

Given its limited size, INbreast poses some challenges in generalization studies of deep learning models. However, many other studies have used it, which makes it easier to compare and reproduce the results achieved. Figs.~\ref{imgDatasetSamples}~(a-b) show samples from the INbreast, in Figs.~\ref{imgDatasetSamples}~(e-f) its ground truth annotation, followed by the mask overlaid on the image, Figs.~\ref{imgDatasetSamples}~(i-j).

More recently, other FFDM datasets were published. DMID~\cite{oza2024-DMID-Dataset} has been introduced to provide additional variability and evaluation opportunities. It is composed of 510 FFDM images, acquired with the \emph{Siemens} equipment \emph{MAMMOMAT 3000 Nova}. Similar to INbreast, this dataset contains 269 annotated cases with abnormalities and 241 non-annotated normal cases. Two images of the DMID dataset are shown in Figs.~\ref{imgDatasetSamples}~(c-d), followed by their ground truth annotation in Figs.~\ref{imgDatasetSamples}~(g-h) and the composition of mask and original image, Figs.~\ref{imgDatasetSamples}~(k-l). After an initial analysis of this dataset, we observed that 269 images had annotations, and the remaining 241 images had no annotations and were considered normal cases.

Existing studies rarely investigate how models, particularly lightweight ones, generalize across datasets with different acquisition characteristics, which are common in real-world scenarios. Therefore, all the 410 images from INbreast were used to train the models, and the 510 images from DMID were used for a final evaluation of the best model.

\subsection{Segmentation Framework}
\label{secSegmentationFramework}

\begin{figure}[H]
     \centering
     \begin{subfigure}[b]{1\textwidth}
         \centering
         \includegraphics[scale=0.55]{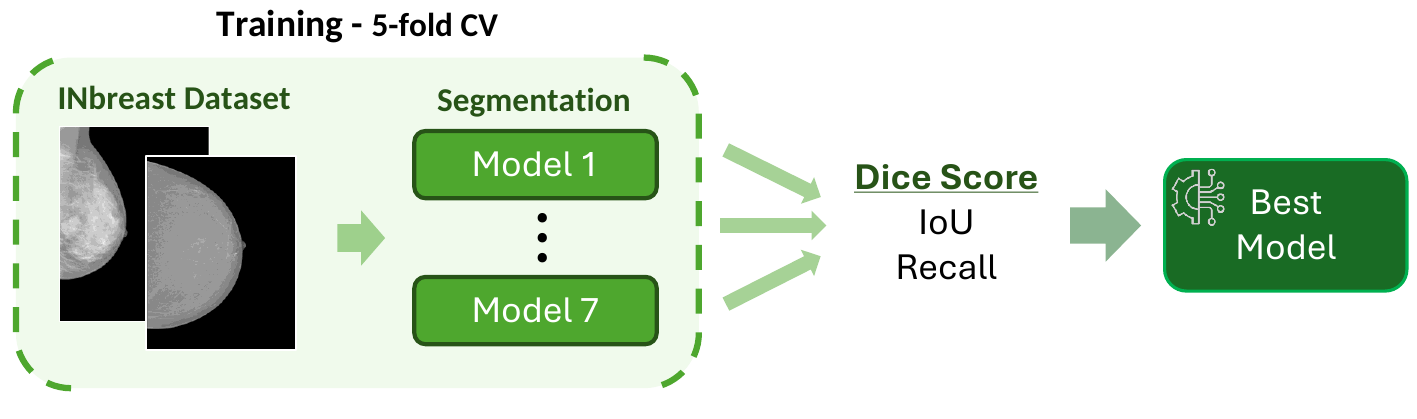}
         \caption{Training step.}
         \label{imgFrameworkTraining}
     \end{subfigure}
     \\
     \begin{subfigure}[b]{1\textwidth}
         \centering
         \includegraphics[scale=0.52]{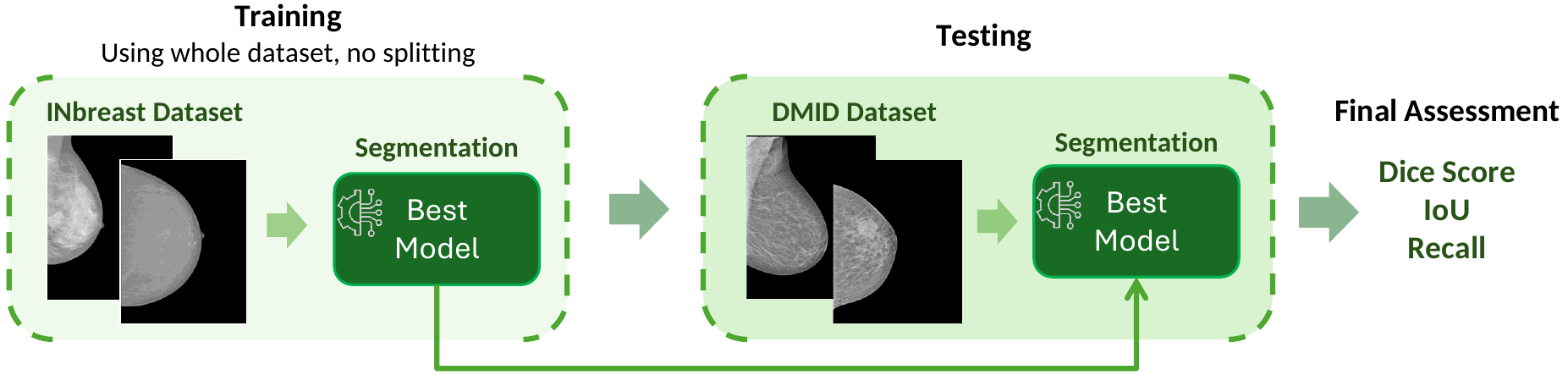}
         \caption{Testing step.}
         \label{imgFrameworkTesting}
     \end{subfigure}
        \caption{Framework established for the models' evaluation and testing.}
        \label{imgFramework}
\end{figure}

This study evaluates the performance and efficiency of multiple deep learning architectures for mammographic lesion segmentation, with a focus on lightweight models suitable for deployment in resource-constrained environments. To accomplish that, the framework shown in Fig.~\ref{imgFramework} was established.

Starting with Fig.~\ref{imgFrameworkTraining}, we considered the INbreast dataset as input to evaluate all seven models and selected the one that performed the best segmentation. For this purpose, each model was independently trained using 5-fold cross-validation (CV). The purpose of using the 5-fold CV scheme is to ensure robust performance estimation and avoid overfitting.

With the 5-fold CV, the dataset is randomly partitioned into five non-overlapping subsets, with each subset used once as a validation set while the remaining subsets are used for training. Each fold runs for 50 epochs, and at the end of an epoch, the best Dice score (a measure of overlap between predicted and ground truth masks) was recorded, followed by the IoU and Recall of the same epoch. These evaluation metrics are formally defined in Sec.~\ref{secEvaluationMetrics}. At the end of a model's training, i.e., at the end of all five folds, we have the top five Dice scores, one per fold. The final model selection, the \emph{best model}, was based on the average Dice coefficient across all the 5-folds. 

After the selection of the best model (end of Fig.~\ref{imgFrameworkTraining}), it was then trained with the entire INbreast dataset without splitting, as shown in Fig.~\ref{imgFrameworkTesting}. Note that the model was not retrained in this last step. The only transfer learning used is described in Sec.~\ref{secImplementationDetails}. The rationale behind this process is that since the INbreast is a relatively small dataset, we wanted to maximize the use of available annotated data, enabling the model to learn from all samples available.

To assess generalization, the selected model was evaluated on the DMID dataset, which contains images acquired under different conditions~\cite{oza2024-DMID-Dataset}. This step, shown on the right-hand side of Fig.~\ref{imgFrameworkTesting}, is critical for evaluating robustness, as models trained on a single dataset may not generalize well due to domain shift and dataset-specific biases~\cite{zech2018}. Moreover, the use of a different dataset mimics a real-world scenario, where different equipment is usually used. Similarly to the training, the Dice score, IoU, and Recall were calculated from each image by comparing the predicted mask against the ground truth. The results will be discussed later in Sec.~\ref{secResults}.

\subsubsection{Implementation Details}
\label{secImplementationDetails}

All models were implemented in the PyTorch library~\cite{NEURIPS2019_PyTroch} and trained on a CPU-only environment to enforce a realistic deployment constraint representative of environments where GPU acceleration, a powerful computer, or cloud access are unavailable. After the model architecture was created, to avoid a cold start and facilitate convergence, the ImageNet-pretrained weights were used for the encoder backbones. This setup defines a concrete operational constraint under which all models are evaluated, enabling a consistent comparison of their efficiency and deployment feasibility.

Input images were processed using methods reported in the literature for medical image processing in deep learning~\cite{oyelade2021a}. Before each training or testing, images were resized to a fixed resolution of 1024$ \times $1024 pixels, normalized, and grouped in batches of 4 images. The batch size limit was chosen to consider the memory constraints of CPU-only systems.

Contrast Limited Adaptive Histogram Equalization (CLAHE) was applied as a stochastic data augmentation technique during training ($p=0.5$) to improve robustness to local contrast variations. CLAHE was not applied during validation, allowing evaluation on the original image distribution. Other data augmentation techniques applied to improve generalization include flipping (horizontal and vertical) and rotation.

Adam optimizer was used with an initial Learning Rate (LR) of $10^{-4}$. To allow the LR automatic update during the training, the \emph{Reduce LR on Plateau} with a threshold of $10^{-3}$ and a minimum of $10^{-6}$ was used. The loss function chosen was a combination of Binary Cross-Entropy and Dice score, both of which will be discussed in Sec.~\ref{secLossFunctions}.

\subsubsection{Selected Lightweight Segmentation Models}
%
%

The models evaluated in this study were selected to represent a spectrum of computational complexity, ranging from standard architectures to highly efficient networks designed for real-time applications.

To fairly compare the models, \emph{U-Net} was trained and used as a baseline owing to its widespread adoption in medical image segmentation. The \emph{U-Net}'s encoder architecture used was the \textit{RestNet34}.

Lightweight alternatives include encoder-decoder variants based on \emph{MobileNetV2}~\cite{sandler2018mobilenetv2} and \emph{EfficientNet}~\cite{tan2019efficientnet}, as well as efficient segmentation architectures such as \emph{FPN}~\cite{lin2017fpn} and \emph{Fast-SCNN}~\cite{poudel2019fastscnn}.

Additionally, attention mechanisms such as Concurrent Spatial and Squeeze \& Excitation (SCSE)~\cite{roy2018scse} blocks were incorporated into selected models to evaluate whether lightweight attention modules can improve segmentation performance without significantly increasing computational cost.

Table~\ref{tblModels} summarizes the approximate number of parameters and Floating-point Operations (FLOPs) of each model. FLOPs were used as a Computational complexity measure considering an input tensor of size $1 \times 1024 \times 1024$~\cite{Dong2022FLOPs}.

\begin{table}[h!]
    \centering
    \caption{List of the lightweight models and their complexity considering an input tensor of $1 \times 1024 \times 1024$.} 
    \label{tblModels}
    \begin{tabular}{l|c|c} 
        \hline\hline
        \rule[-1ex]{0pt}{3.5ex} \textbf{Model} & \textbf{Params (M)} & \textbf{FLOPs (G)} \\
        \hline\hline
        \rule[-1ex]{0pt}{3.5ex} U-Net (ResNet34) & 24.43 & 248.09 \\
        \hline
        \rule[-1ex]{0pt}{3.5ex} MobileNet V2 & 6.63 & 108.21 \\
        \hline
        \rule[-1ex]{0pt}{3.5ex} MobileNet V2 $+$ SCSE & 6.89 & 108.66 \\
        \hline
        \rule[-1ex]{0pt}{3.5ex} FPN (ResNet18) & 13.04 & 139.51 \\
        \hline
        \rule[-1ex]{0pt}{3.5ex} Fast-SCNN & 2.26 & 10.52 \\
        \hline
        \rule[-1ex]{0pt}{3.5ex} EfficientNet Lite & 5.61 & 95.23 \\
        \hline
        \rule[-1ex]{0pt}{3.5ex} EfficientNet Lite  $+$ SCSE & 5.67 & 95.68 \\
        \hline
    \end{tabular}
\end{table} 

\emph{MobileNetV2} was evaluated in two versions, one with and another without the SCSE block. This slightly increases the number of parameters from 6.63M to 6.89M. A similar approach was applied to the \emph{EfficientNet}'s \textit{Lite}~\cite{tan2019efficientnet} variant. The \emph{FPN}~\cite{lin2017fpn} variant evaluated has \textit{ResNet18}~\cite{he2016resnet} as backbone. Lastly, the \emph{Fast-SCNN}~\cite{poudel2019fastscnn} has the \emph{MobileNetV2} encoder.

Since mammographic images are grayscale images, the input of each model has only one channel as opposed to natural images, which usually have 3 (RGB) channels. Moreover, to improve model generalization, transfer learning was performed by using weights of the pre-trained models on ImageNet. Although ImageNet is composed of natural images, it has been demonstrated that even for medical images, transfer learning has improved the results~\cite{shin2016}.

\subsubsection{Loss Functions}
\label{secLossFunctions}

Mammographic images, such as the ones shown in Fig.~\ref{imgDatasetSamples}, have a significant class imbalance, where the amount of background pixels is much higher compared to the ones representing the breast as well as its structures of interest. 

Due to this imbalance in mammographic lesion segmentation, particularly for micro-structures occupying less than $1\%$ of the image area, the optimization function is a combination of Dice Score and \emph{Binary Cross-Entropy} (BCE)~\cite{taghanaki2019comboloss}. 

The \emph{Binary Cross-Entropy} is defined as:
\begin{equation}
\label{eqBCELoss}
    BCE_{Loss} = -\displaystyle\sum_i \left[ g_i \log(g_i) + (1 - g_i) \log(1 - p_i) \right],
\end{equation}

\noindent where $p_i$ and $g_i$ correspond to the pixel $i$ in the predicted mask and in the ground truth, respectively. The term $(1 - g_i) \log(1 - p_i)$ is zero when the prediction is correct, thus it tends to penalize false-positives.


The Dice score is widely used as a metric to evaluate the performance of a segmentation model, and this version will be explained in  Sec.~\ref{secEvaluationMetrics}. A version of the squared Dice score as a loss function has been proposed, leveraging its power~\cite {milletari2016vnet}. Therefore, the $Dice_{Loss}$ is defined as:
\begin{equation}
\label{eqDICELoss}
    Dice_{Loss} = 1 - \dfrac{2 \displaystyle\sum_i p_i g_i + \epsilon}{\displaystyle\sum_i p^2_i + \displaystyle\sum_i g^2_i  + \epsilon},
\end{equation}
\noindent where $p_i$ and $g_i$ correspond to the pixel $i$ in the predicted mask and in the ground truth, respectively, and $\epsilon$ is a stability factor to avoid division by 0, mainly when both $p$ and $g$ are empty.

Therefore, the \textit{Loss} function was defined as a combination of both $BCE_{Loss}$ (Eq.~\ref{eqBCELoss}) and $Dice_{Loss}$ (Eq.~\ref{eqDICELoss}):
\begin{equation}
\label{eqLoss}
    Loss = BCE_{Loss} + Dice_{Loss}
\end{equation}

\subsection{Evaluation Metrics}
\label{secEvaluationMetrics}
When evaluating medical image segmentation models, one must take into consideration that the classes are highly imbalanced. For example, the background is much bigger than the foreground in a proportion of 9 to 1. For that reason, metrics that have similar weight for true positive and true negative, such as accuracy and specificity, tend to lead to misinterpretation~\cite{muller_towards_2022}.

Metrics better suited for segmentation are the ones that focus on true positives without considering the true negative cases. Thus, besides the aforementioned considerations, the metrics chosen for models' performance evaluation were also selected based on their widespread use in medical image segmentation and their ability to capture complementary aspects of performance, including overlap accuracy (Dice, IoU) and sensitivity to lesion detection (Recall)~\cite{taha_metrics_2015}.

Dice Score (DSC), also known as Dice Similarity Coefficient or F1-score, is the harmonic mean between sensitivity and precision. The Dice Score is defined as:
\begin{equation}
\label{eq:dice_bool}
Dice = \frac{2 TP}{2TP + FP + FN},
\end{equation}
\noindent where TP stands for \emph{True Positive}, FP \emph{False Positive} and, FN \emph{False Negative}. 

Another measure considered for the models' evaluation in this study is the Intersection over Union (IoU). Similar to the Dice score, IoU analyzes the overlap between the predicted mask and the ground truth. Therefore, IoU is defined as:
\begin{equation}
    IoU = \frac{TP}{TP + FP + FN}
\end{equation}

To evaluate the true positive rate, the Recall measure was used~\cite{taha_metrics_2015}: 
\begin{equation}
    \text{Recall} = \frac{TP}{TP + FN}
\end{equation}

Although the Dice score and IoU are mathematically related and both quantify spatial overlap, they provide complementary perspectives in practice. Dice is more sensitive to small structures and is widely used in medical image segmentation, particularly under class imbalance conditions. In contrast, IoU penalizes false positives and false negatives more strictly, offering a more conservative estimate of segmentation performance. Meanwhile, recall focuses on the true positive rate, i.e., the proportion of ground truth pixels correctly identified as positive by the model. Reporting these metrics ensures consistency with prior work and facilitates a more comprehensive evaluation of model behavior.

\section{Results and Discussion}
\label{secResults}
This section presents a comparative analysis of segmentation performance across the evaluated models, considering both accuracy metrics (Dice, IoU, Recall) and computational efficiency (model size and FLOPs). Cross-dataset generalization results are also discussed.

From a deployment perspective, the results suggest that enforcing computational constraints (e.g., CPU-only environment) does not necessarily prevent achieving useful segmentation performance.

\subsection{Quantitative Results}
\label{secQuantitativeRes}

\begin{figure}[htb]
    \centering
    \includegraphics[scale=0.95]{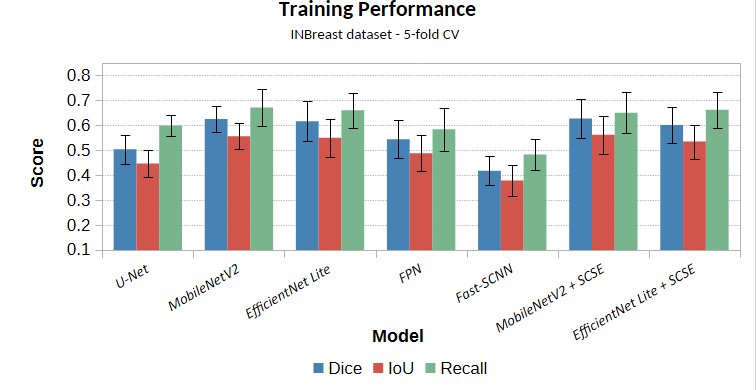}
    \caption{Training results based on INbreast dataset.}
    \label{figTrainingResults}
\end{figure}

As discussed in Sec.~\ref{secSegmentationFramework}, the initial experiment aimed to identify the best model among the ones evaluated. Table~\ref{tblTrainingResults} and the graph in Fig.~\ref{figTrainingResults} present the segmentation performance of all evaluated models across the INbreast dataset using 5-fold cross-validation. It is worth mentioning that the \textit{U-Net} was included as a baseline.

\begin{table}[h!]
    \caption{Segmentation performance and model complexity comparison. The best model was selected based on the Dice score.}
    \label{tblTrainingResults}
    \centering
    \begin{tabular}{l|c|c|c|c}
        \hline\hline
        \rule[-1ex]{0pt}{3.5ex} \textbf{Model} & \textbf{Dice Score} $\uparrow$ & \textbf{IoU} $\uparrow$ & \textbf{Recall} $\uparrow$ & \textbf{Params (M)} $\downarrow$ \\
        \hline\hline
        \rule[-1ex]{0pt}{3.5ex} U-Net (ResNet34) & 0.4532 $\pm$ 0.0569 & 0.3962 $\pm$ 0.0537 & 0.5486 $\pm$ 0.0407 & 24.43 \\
        \rule[-1ex]{0pt}{3.5ex} MobileNetV2 & 0.5751 $\pm$ 0.0504 & 0.5057 $\pm$ 0.0515 &  0.6206 $\pm$ 0.0728 & 6.63 \\
        \rule[-1ex]{0pt}{3.5ex} \textbf{MobileNetV2 + SCSE} & \textbf{0.5766 $\pm$ 0.0780} & \textbf{0.5116 $\pm$ 0.0752} & \textbf{0.5998 $\pm$ 0.0823} & \textbf{6.89} \\
        \rule[-1ex]{0pt}{3.5ex} FPN (ResNet18) & 0.4938 $\pm$ 0.0765 & 0.4378 $\pm$ 0.0721 & 0.5335 $\pm$ 0.0860 & 13.04 \\
        \rule[-1ex]{0pt}{3.5ex} Fast-SCNN & 0.3675 $\pm$ 0.0583 &  0.3283 $\pm$ 0.0630 & 0.4326 $\pm$ 0.0620 & 2.26 \\
        \rule[-1ex]{0pt}{3.5ex} EfficientNet Lite & 0.5655 $\pm$ 0.0797 &  0.4998 $\pm$ 0.0758 & 0.6101 $\pm$ 0.0702 & 5.61 \\
        \rule[-1ex]{0pt}{3.5ex} EfficientNet Lite + SCSE & 0.5510 $\pm$ 0.0722 & 0.4846 $\pm$ 0.0681 & 0.6116 $\pm$ 0.0711 & 5.67 \\
        \hline\hline
    \end{tabular}
\end{table}

The decision on the best model was based solely on the average Dice score across all 5 folds. Notably, \textit{MobileNetV2}-based architectures achieved the highest Dice scores of $0.5751$ and $0.5766$, respectively. Compared to \textit{U-Net}, both \textit{MobileNetV2} and \textit{MobileNetV2 + SCSE} have approximately 75\% fewer parameters and consequently require fewer FLOPs (see Table~\ref{tblModels}), demonstrating that comparable segmentation performance can be achieved under strict computational constraints, with substantially reduced model complexity.

These results indicate that lightweight models occupy a favorable region of the accuracy--efficiency spectrum, achieving competitive performance with significantly reduced computational requirements.

The difference in Dice score between the two \textit{MobileNetV2} models is $0.0015$, with the model equipped with the Concurrent Spatial and Squeeze \& Excitation (SCSE) module achieving the highest score. According to the IoU, this difference is slightly higher, of $0.0059$, but might not be significant when considering the standard deviation. On the other hand, the \textit{MobileNetV2} reached a Recall (True Positive Rate) of $0.6206$, the highest among all the models evaluated.

Another model that achieved remarkable performance, considering the constraints of this study, was the \textit{EfficientNet Lite}. Both versions, with and without the SCSE block, are much smaller than \textit{U-Net} (also 75\% fewer parameters) and have 1 M fewer parameters than \textit{MobileNetV2}, and they were capable of performing well in training with a Dice score of $0.5510$ and $0.5655$, respectively. As opposed to what was seen in \textit{MobileNetV2}, the \textit{EfficientNet Lite} model with SCSE produced a lower Dice score compared to the other version without it.

Other models, such as \textit{FPN} and \textit{Fast-SCNN}, had lower scores compared to \textit{MobileNetV2} and \textit{EfficientNet Lite}. \textit{FPN} achieved an average Dice score of $0.4938$, outperforming the \textit{U-Net} baseline while maintaining a lower computational cost. \textit{Fast-SCNN}, as can be observed in the graph of Fig.~\ref{figTrainingResults}, had the lowest results compared to the other models. This might be related to the model being the smallest, with only 2.26 M parameters. This is approximately 50\% less than the next bigger model, \textit{EfficientNet Lite}, and 90\% smaller than \textit{U-Net}.

\begin{figure}[!htb]
    \centering
    \includegraphics[scale=0.7]{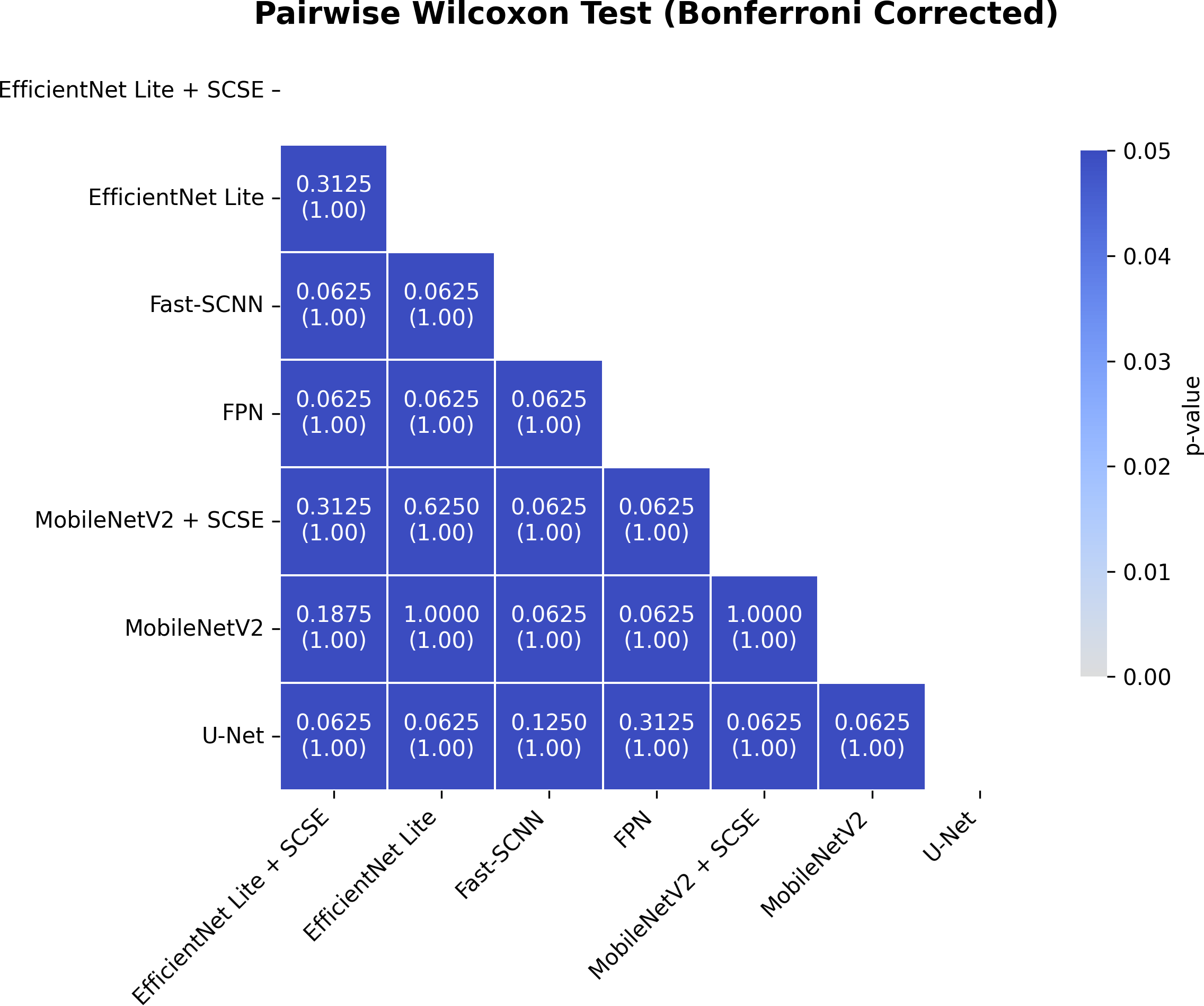}
    \caption{Wilcoxon signed-rank $p-values$ across the five cross-validation folds. In parentheses is the result of Bonferroni correction.}
    \label{figWilcoxonPlot}
\end{figure}

To assess whether the observed differences between models are statistically significant, pairwise Wilcoxon signed-rank tests were performed across the five cross-validation folds. Bonferroni correction was applied to account for multiple comparisons. As shown in Fig.~\ref{figWilcoxonPlot}, each box has the $p-value$ of the pair-wise comparison with the Bonferroni correction result inside the parentheses. As one can note, besides the results discussed earlier, no statistically significant differences were observed between models ($p > 0.05$), indicating that performance differences are not statistically robust under the current experimental setup.

\subsection{Cross-Dataset Generalization}
In clinical settings, it is common for physicians to analyze images acquired with different equipment. Therefore, given the various sources of data, CAD systems must be robust enough to generalize across different characteristics. 

To assess the generalization capability of the previously selected best-performing model (\textit{MobileNetV2 + SCSE}) across different datasets, further evaluation was performed on the DMID dataset. The Table~\ref{tab:generalization} shows the Dice score, IoU, and Recall considering a segmentation threshold of $0.5$. 

\begin{table}[h!]
    \centering
    \caption{Cross-dataset generalization results (trained on INbreast, tested on DMID) considering a threshold of $0.5$.}
    \label{tab:generalization}
    \begin{tabular}{l|c|c|c}
        \hline\hline
        \textbf{Model} & \textbf{Dice Score} $\uparrow$ & \textbf{IoU} $\uparrow$ & \textbf{Recall} $\uparrow$ \\
        \hline\hline
        MobileNetV2 + SCSE & 0.4381 & 0.4102 & 0.5737 \\
        \hline
    \end{tabular}
\end{table}

The observed performance drop highlights the challenges associated with domain shift, which has been widely reported in medical imaging applications~\cite{zech2018}. 
It is important to note that the Dice score decreased by 24.02\%, accompanied by a 19.82\% reduction in IoU when evaluated on the unseen dataset, indicating a noticeable drop in segmentation accuracy. However, the Recall exhibited only a modest decrease of 4.35\%, suggesting that the model preserved its ability to identify most of the relevant regions. 

This behavior indicates that, while the model retains sensitivity to target structures, its precision and boundary delineation are affected under domain shift. Nevertheless, the results still point to a degree of cross-dataset generalization, particularly considering the use of a lightweight architecture.

\begin{figure}[!h]
     \centering
     \begin{subfigure}[b]{0.18\textwidth}
         \centering
         \includegraphics[scale=0.15]{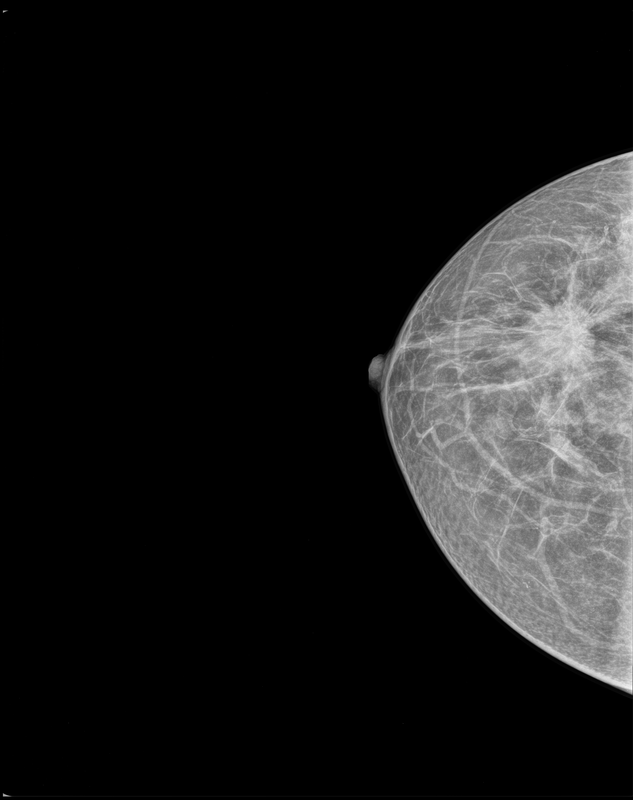}
         \caption{}
     \end{subfigure}
     \hfill
     \begin{subfigure}[b]{0.18\textwidth}
         \centering
         \includegraphics[scale=0.15]{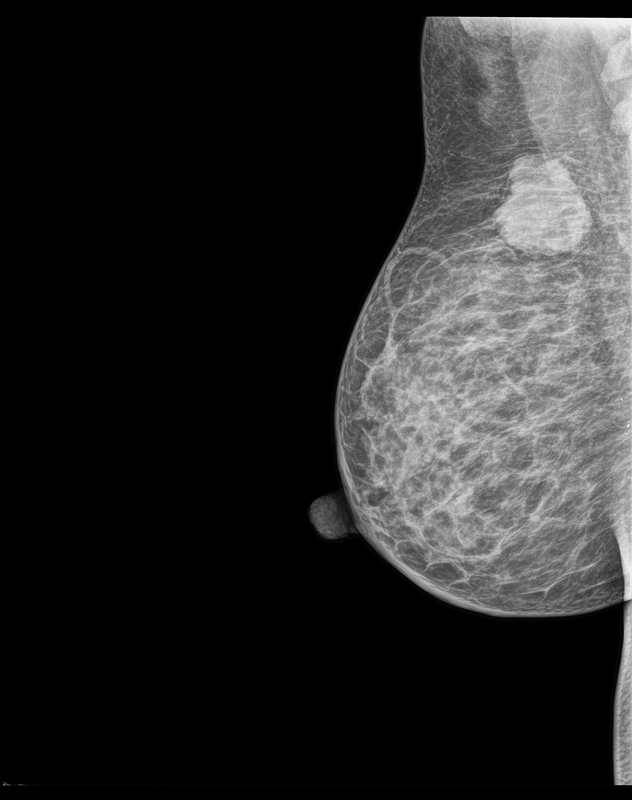}
         \caption{}
     \end{subfigure}
     \hfill
     \begin{subfigure}[b]{0.18\textwidth}
         \centering
         \includegraphics[scale=0.15]{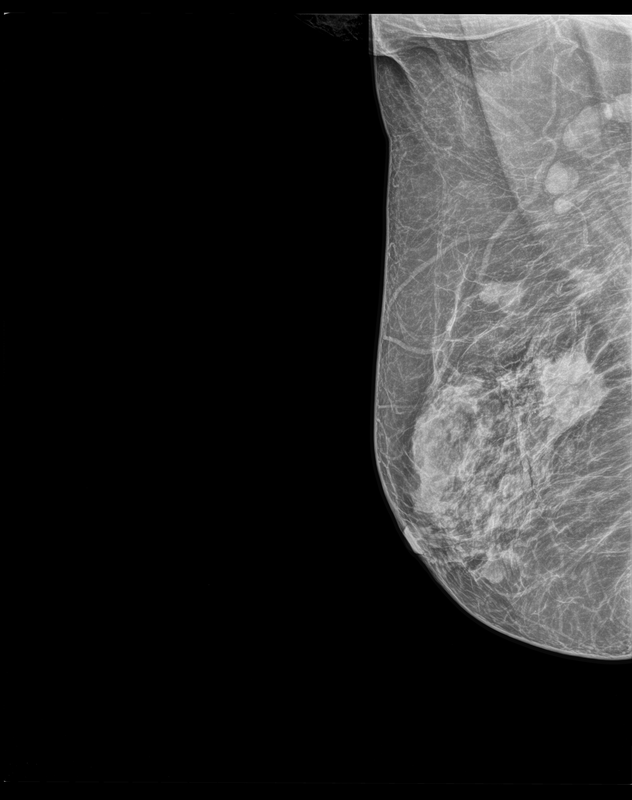}
         \caption{}
     \end{subfigure}
     \hfill
     \begin{subfigure}[b]{0.18\textwidth}
         \centering
         \includegraphics[scale=0.15]{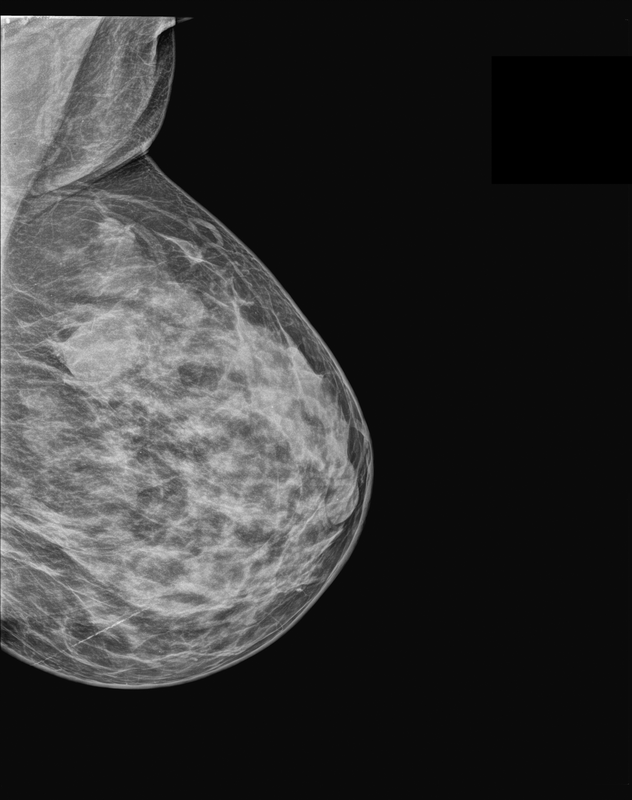}
         \caption{}
     \end{subfigure}
     \\
      \begin{subfigure}[b]{0.18\textwidth}
         \centering
         \includegraphics[scale=0.15]{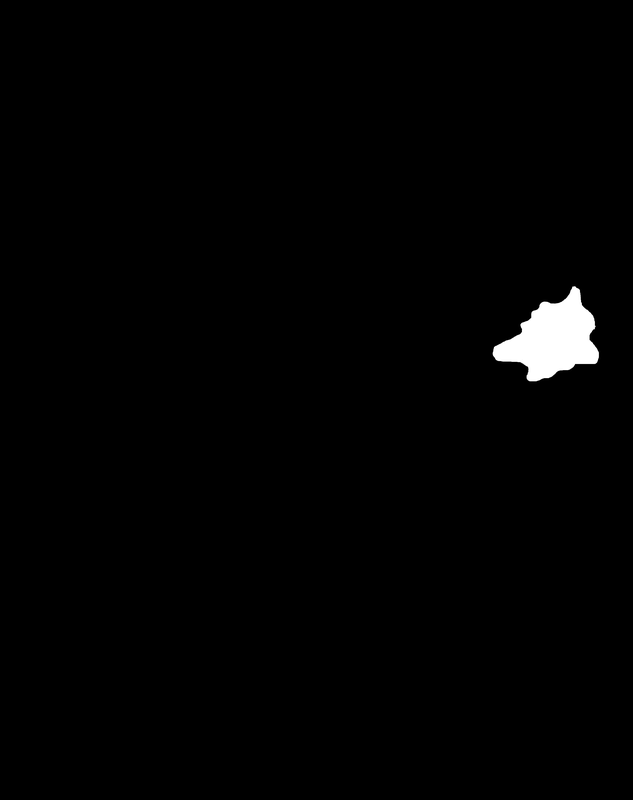}
         \caption{}
     \end{subfigure}
     \hfill
     \begin{subfigure}[b]{0.18\textwidth}
         \centering
         \includegraphics[scale=0.15]{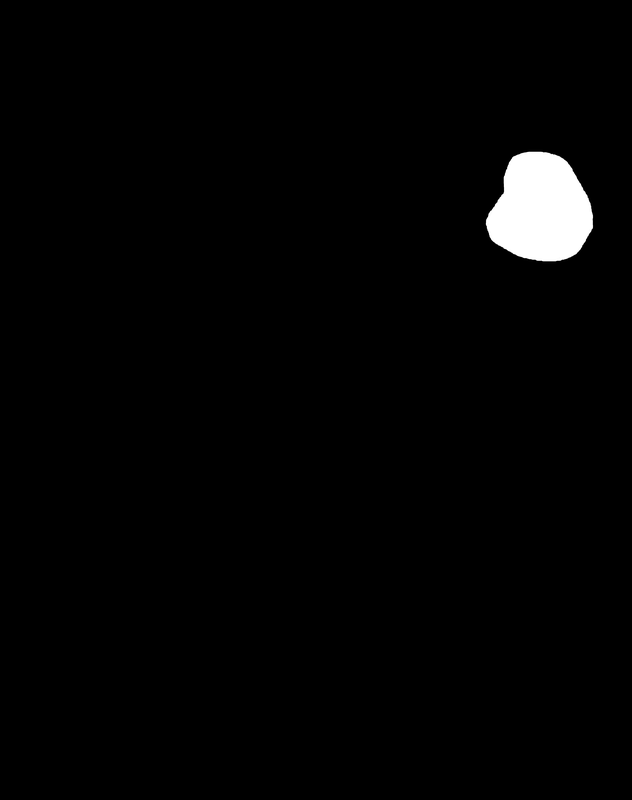}
         \caption{}
     \end{subfigure}
     \hfill
     \begin{subfigure}[b]{0.18\textwidth}
         \centering
         \includegraphics[scale=0.15]{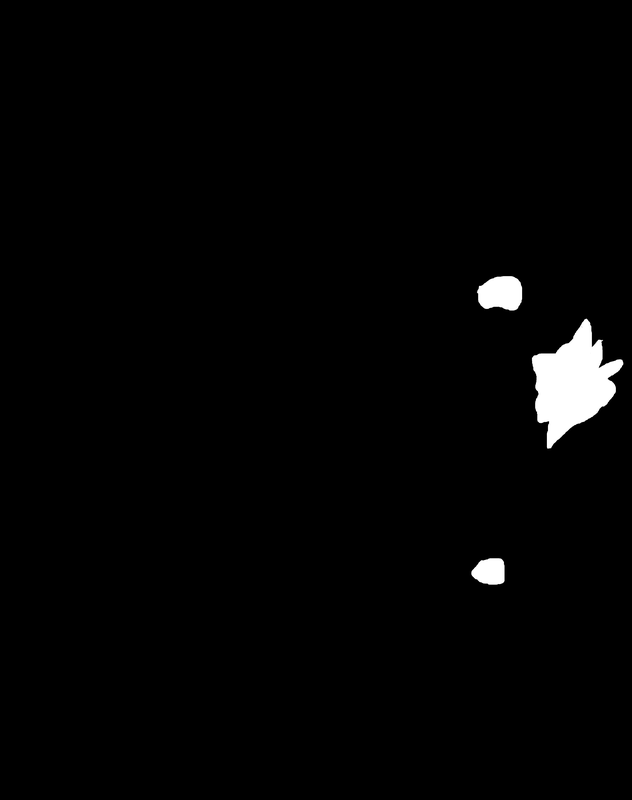}
         \caption{}
     \end{subfigure}
     \hfill
     \begin{subfigure}[b]{0.18\textwidth}
         \centering
         \includegraphics[scale=0.15]{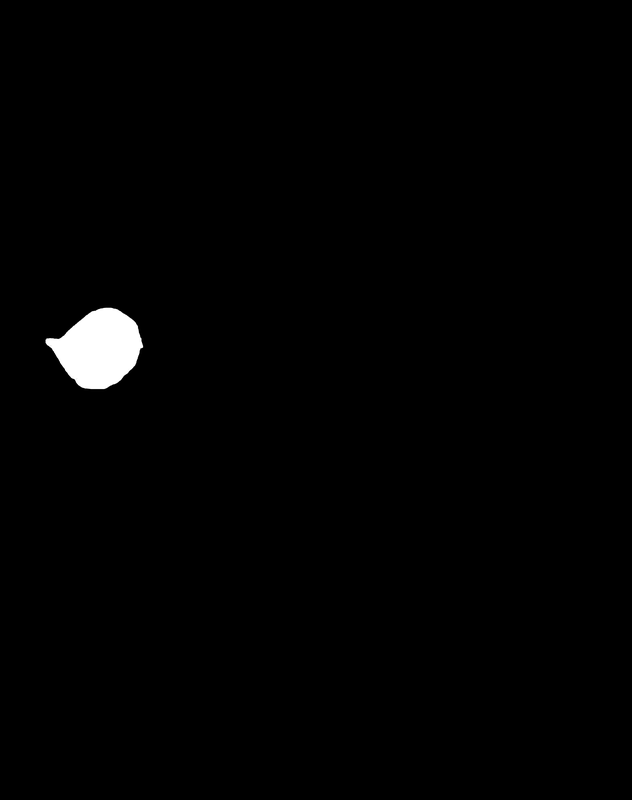}
         \caption{}
     \end{subfigure}
     \\
      \begin{subfigure}[b]{0.18\textwidth}
         \centering
         \includegraphics[scale=0.15]{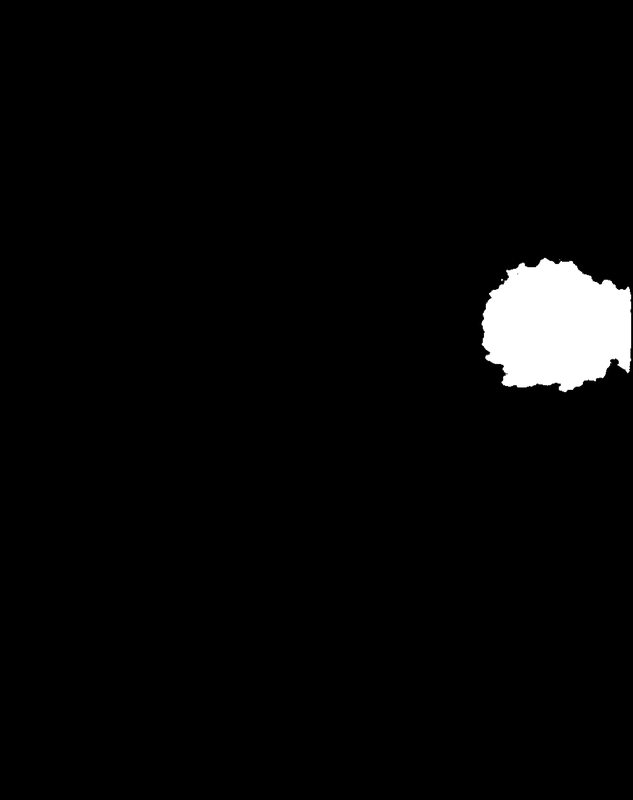}
         \caption{}
     \end{subfigure}
     \hfill
     \begin{subfigure}[b]{0.18\textwidth}
         \centering
         \includegraphics[scale=0.15]{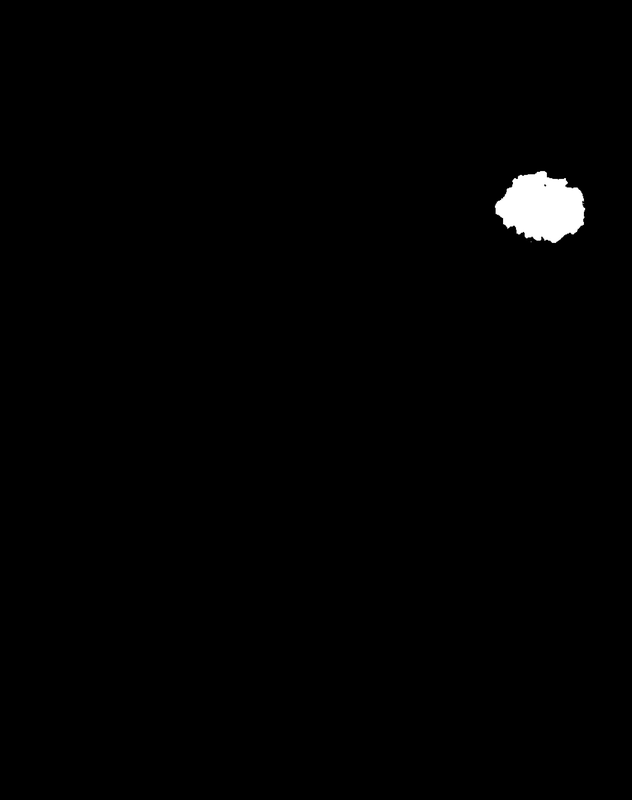}
         \caption{}
     \end{subfigure}
     \hfill
     \begin{subfigure}[b]{0.18\textwidth}
         \centering
         \includegraphics[scale=0.15]{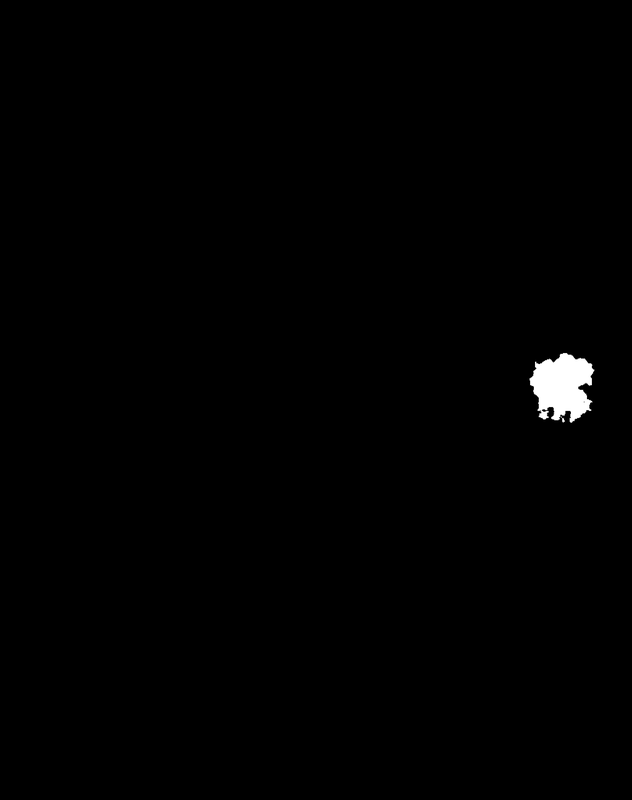}
         \caption{}
     \end{subfigure}
     \hfill
     \begin{subfigure}[b]{0.18\textwidth}
         \centering
         \includegraphics[scale=0.15]{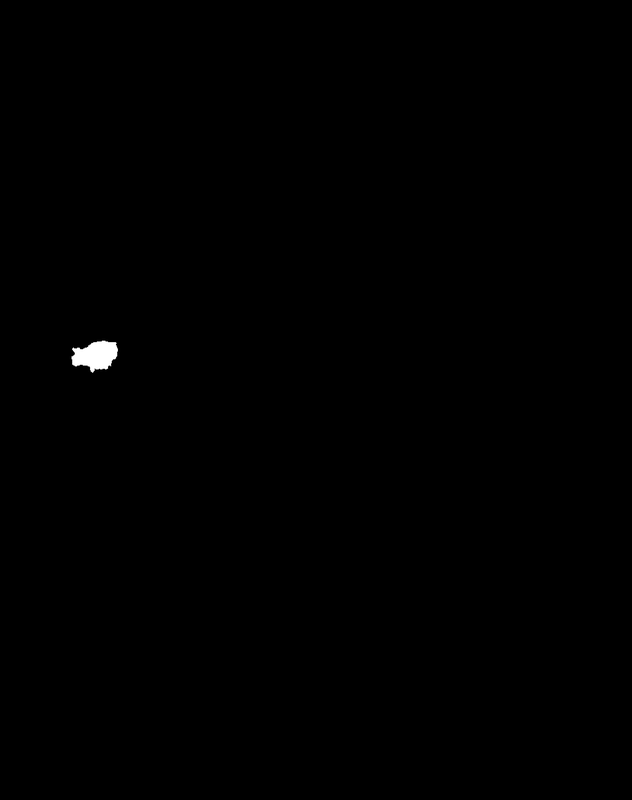}
         \caption{}
     \end{subfigure}
     \\
      \begin{subfigure}[b]{0.18\textwidth}
         \centering
         \includegraphics[scale=0.15]{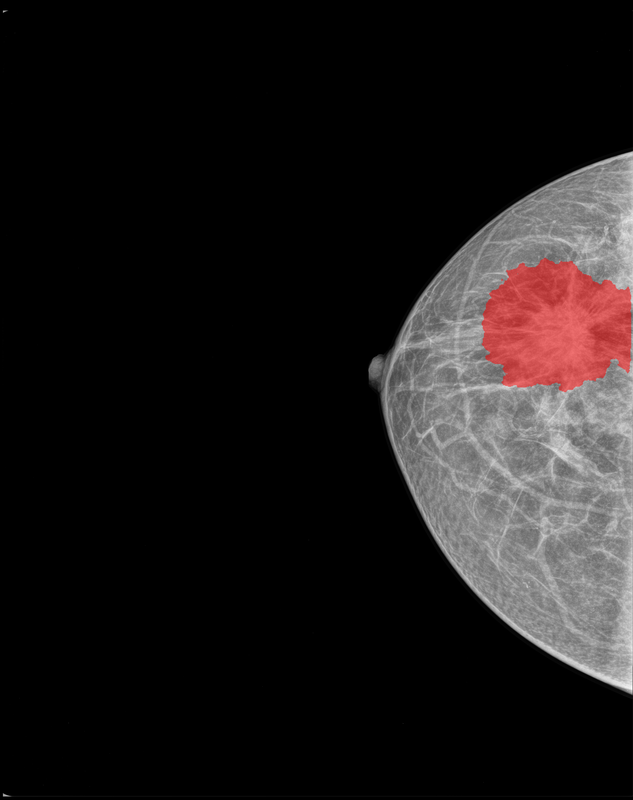}
         \caption{}
     \end{subfigure}
     \hfill
     \begin{subfigure}[b]{0.18\textwidth}
         \centering
         \includegraphics[scale=0.15]{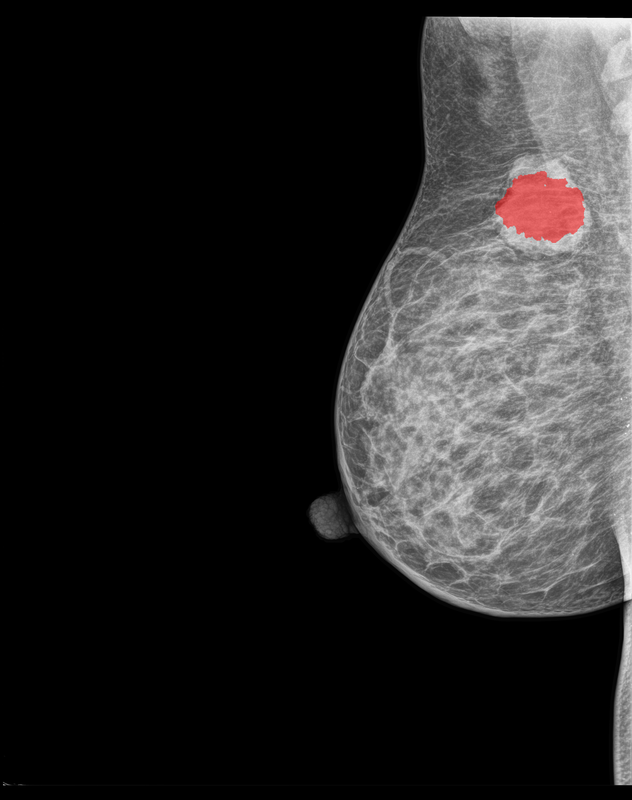}
         \caption{}
     \end{subfigure}
     \hfill
     \begin{subfigure}[b]{0.18\textwidth}
         \centering
         \includegraphics[scale=0.15]{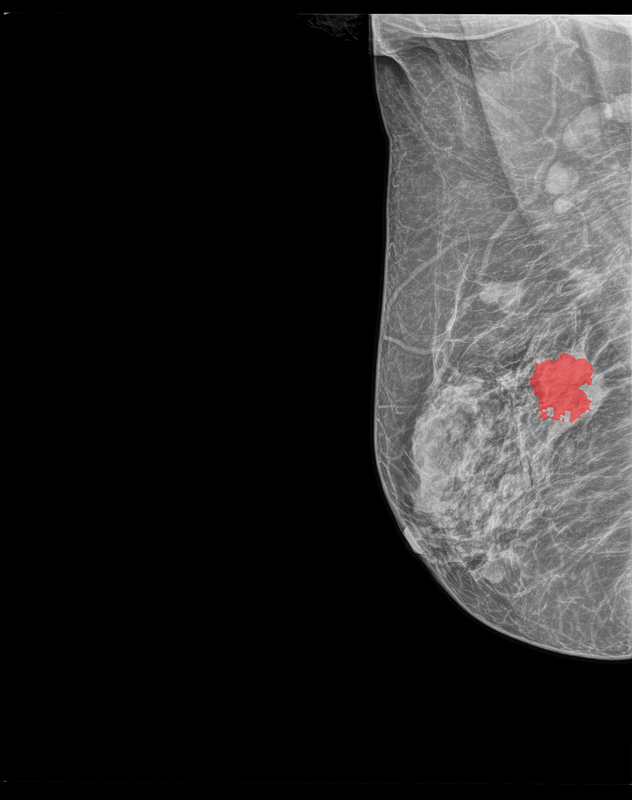}
         \caption{}
     \end{subfigure}
     \hfill
     \begin{subfigure}[b]{0.18\textwidth}
         \centering
         \includegraphics[scale=0.15]{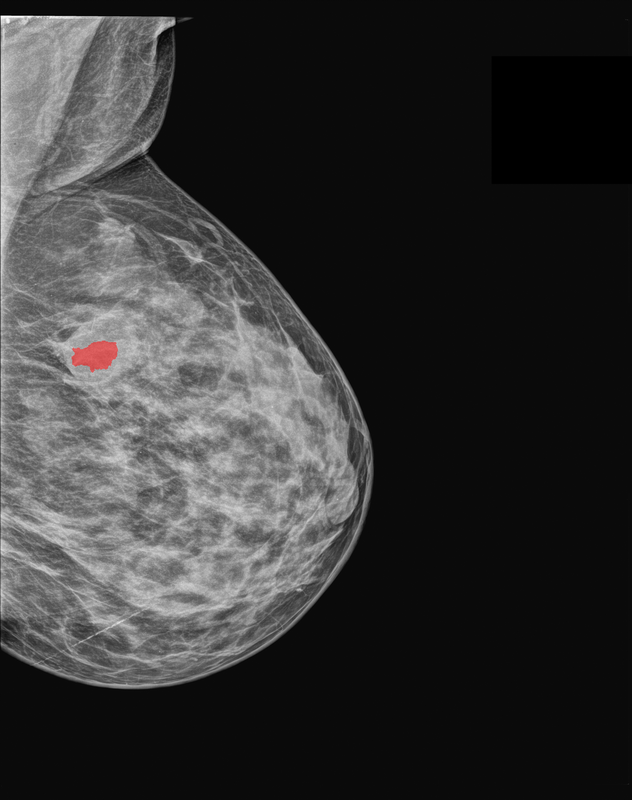}
         \caption{}
     \end{subfigure}
        \caption{Images from DMID dataset. (a-d) original images; (e-h) ground truth masks; (i-l) predicted masks at a threshold of 0.5, and (m-p) predicted masks overlaid on the images.}
        \label{imgTestingResults}
\end{figure}

For visual inspection, in Fig.~\ref{imgTestingResults} we show four samples of the DMID dataset: The original images are shown in Fig.~\ref{imgTestingResults} (a-d); followed by their ground truth masks in Fig.~\ref{imgTestingResults} (e-h), the masks predicted by \textit{MobileNetV2 + SCSE} model at a threshold of $0.5$ in Fig.~\ref{imgTestingResults} (i-l), and the predicted mask overlaid on the original images in Fig.~\ref{imgTestingResults} (m-p).

As can be seen in the first column of Fig.~\ref{imgTestingResults} (image \verb|ID:015|), some of the regions (Fig.~\ref{imgTestingResults}~(i)) are over segmented by the model compared to the ground truth producing a Recall of 1.0. For that case, the accuracy in terms of Dice score and IoU were $0.5706$ and $0.3992$, respectively. It is interesting to note that, compared to the image in the second column, the first column's image shows a much subtle lesion, nevertheless, the model did a good segmentation.

The second case Fig.~\ref{imgTestingResults}~(b) (image \verb|ID:315|), clearly show a mass that was under-segmented. The Dice score, IoU and Recall in this case were, $0.6202$, $0.4495$ and, $0.4495$, respectively. Besides the mass being visible, the segmentation went reasonably well by partially flagging (see predicted mask in Fig.~\ref{imgTestingResults}~(j)) the suspicious region.

Fig.~\ref{imgTestingResults}~(c) (image \verb|ID:446|) shows a mammogram with two three suspicious spots, according to the ground truth mask (Fig.~\ref{imgTestingResults}~(g)). For this case, the model did not perform a good segmentation by missing two out of three lesions. The Dice score, IoU and Recall reached were $0.4627$, $0.3009$ and, $0.3333$, respectively.

In challenging cases such as in Fig.~\ref{imgTestingResults}~(d) (image \verb|ID:477|) where the image shows a dense breast, the model under-segmented achieving a Dice score of $0.0479$, IoU and Recall of $0.0245$. Dense breasts are reported in the literature as the most difficult cases since health tissue can mask lesions. Even though, the model was capable of partially predict the suspicious region which might help the radiologist to perform further investigation. 

It is important to remind that the results shown in Fig.~\ref{imgTestingResults} were produced at a segmentation threshold of $0.5$ in a dataset never seen by the model. Next, we further explore the variation of the threshold and its impact into the mask prediction.

\subsection{Threshold Sensitivity Analysis}

To assess the impact of threshold selection on segmentation performance under domain shift, we evaluated the model using thresholds ranging from 0.1 to 0.9. Table~\ref{tab:threshold} summarizes the results in terms of Dice coefficient, Intersection over Union (IoU), and Recall.

\begin{table}[ht]
    \centering
    \caption{Segmentation performance as a function of threshold on the external dataset (DMID). A threshold of 0.5 was used as a reference.}
    \label{tab:threshold}
    \begin{tabular}{c|c|c|c}
        \hline
        \textbf{Threshold} & \textbf{Dice} & \textbf{IoU} & \textbf{Recall} \\
        \hline
        0.1 & 0.3958 & 0.3671 & 0.5835 \\
        0.2 & 0.4104 & 0.3820 & 0.5800 \\
        0.3 & 0.4180 & 0.3898 & 0.5776 \\
        0.4 & 0.4299 & 0.4019 & 0.5756 \\
        \textbf{0.5} & \textbf{0.4381} & \textbf{0.4102} & \textbf{0.5737} \\
        0.6 & 0.4442 & 0.4165 & 0.5718 \\
        0.7 & 0.4617 & 0.4341 & 0.5695 \\
        0.8 & 0.4627 & 0.4355 & 0.5663 \\
        0.9 & 0.4751 & 0.4484 & 0.5608 \\
        \hline
    \end{tabular}
\end{table}

\begin{figure}[!ht]
    \centering
    \includegraphics[scale=0.8]{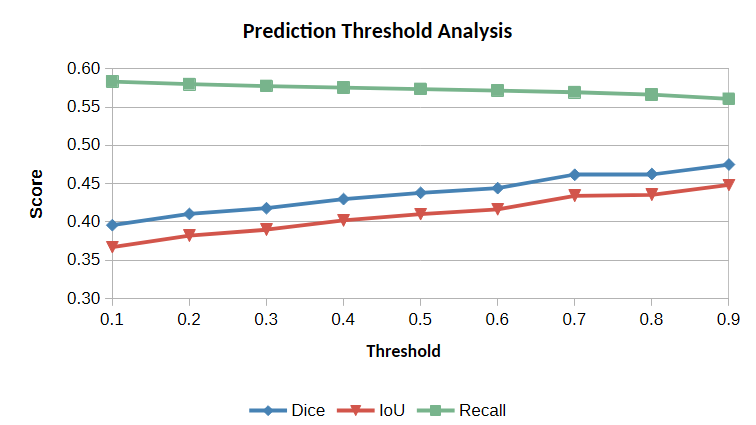}
    \caption{Segmentation performance (Dice, IoU, and Recall) as a function of threshold on the external dataset.}
    \label{fig:threshold_curves}
\end{figure}

As shown in Table~\ref{tab:threshold} and Figure~\ref{fig:threshold_curves}, increasing the threshold results in a slight improvement in Dice and IoU, while recall remains relatively stable across all evaluated values (approximately 0.57). The best performance in terms of Dice and IoU was observed at a threshold of 0.9.

The limited variation in recall across different thresholds suggests that the model's ability to detect lesion regions is largely insensitive to threshold selection. In contrast, the modest improvements in Dice and IoU at higher thresholds indicate a slight reduction in over-segmentation effects.

These findings suggest that the observed performance limitations are not primarily due to suboptimal threshold selection, but rather to underlying differences in data distribution between the training and external datasets. In particular, the relatively stable recall combined with low spatial overlap metrics indicates that the model retains sensitivity to lesion presence, while failing to achieve precise boundary delineation under domain shift conditions.

\section{Conclusions and Future Works}

This study investigates the use of lightweight deep learning models for mammographic lesion segmentation, with an emphasis on computational efficiency and deployability. The results of this work aim to inform the design of practical CAD systems suitable for resource-constrained environments.

The results demonstrate a clear trade-off between segmentation performance and computational efficiency. \textit{U-Net} was evaluated as a baseline model for comparison purposes against 4 other models, namely \textit{MobileNetV2}, \textit{FPN}, \textit{Fast-SCNN}, and \textit{EfficientNet Lite}. \textit{MobileNetV2} and \textit{EfficientNet Lite} were tested in two versions, one with attention (through the SCSE block) and another without it. 

Through a 5-fold cross-validation with the INbreast dataset (410 samples), almost all the models evaluated performed better than the \textit{U-Net}, except \textit{Fast-SCNN}, the smallest one. The best accuracy in terms of Dice score was $0.5766 \pm 0.0780$ achieved by \textit{MobileNetV2} with the SCSE block. This shows that lightweight architectures provide a compelling alternative when computational resources are limited.

A second evaluation aimed at assessing the generalization of the \textit{MobileNetV2}, the best model in terms of Dice score during training. For that, the domain transfer was evaluated using an unseen dataset, the DMID. Besides the expected degradation, \textit{MobileNetV2} achieved results that make it particularly suitable for deployment in real-world CAD systems operating under hardware constraints.

The main limitation of this work is the dataset used for training. The INbreast dataset is widely used in papers not only for segmentation but also for cancer detection. Nevertheless, it has only 410 images with some lesions underrepresented, such as architectural distortion, which has only a few cases. Another limitation is the number of models trained. Thus, our future directions are: first, expand the training dataset in order to reduce the lesions' classes imbalance and, second, evaluate more optimized architectures.

The findings in this paper support the hypothesis that \textbf{lightweight models are capable of achieving competitive accuracy even under domain transfer}. Thus, we argue that increasing model complexity does not necessarily translate into proportionally improved performance under constrained settings, making it possible to use CAD systems in environments with limited computational power, especially without GPUs.

Future work will further formalize this trade-off by incorporating explicit latency and memory measurements, enabling more precise deployment-oriented evaluation.

\section{Code, Data, and Materials Availability}
\label{secDataAvailability}
The code, the best model, and experimental results supporting this study are publicly available at the Open Science Framework (OSF): \url{https://osf.io/dm3vq/}. Due to dataset licensing restrictions, the mammographic images are not redistributed. However, instructions for accessing the datasets are provided in the OSF repository.

\section*{Disclosures}
The author declares no competing financial interests. Helder Oliveira is affiliated with GnosisX, an early-stage initiative involved in the development of medical imaging software, including projects related to this work. This affiliation did not influence the study design, analysis, or interpretation of the results.

\section*{Acknowledgments}
The author acknowledges the use of Grammarly and ChatGPT (OpenAI) for language editing and grammar correction to improve clarity and readability. All scientific content, including methodology, analysis, and interpretation, was developed entirely by the author.


\bibliography{references}   
\bibliographystyle{spiejour}   



\vspace{2ex}\noindent\textbf{Helder Oliveira} is a researcher specializing in medical imaging and data analysis. He holds a PhD in electrical engineering, with a focus on medical image analysis. His work includes breast cancer detection, computer-aided detection (CAD) systems, and machine learning approaches for medical imaging. He is also interested in probabilistic graphical models such as Bayesian networks and Structural Equation Models.



\end{spacing}
\end{document}